\pdfoutput=1
\documentclass[letterpaper, 10 pt, conference]{ieeeconf}  

\IEEEoverridecommandlockouts                              

\usepackage{graphics} 
\DeclareGraphicsExtensions{.pdf,.png,.jpg}

\usepackage{epsfig} 
\usepackage{times} 
\usepackage{amsmath} 
\usepackage{amssymb}  
\usepackage{subfigure}
\usepackage[cjk]{kotex}
\usepackage{cite}
\usepackage{mwe}
\usepackage{xcolor}
\usepackage{url}
\usepackage{multirow}
\usepackage{fixltx2e}
\usepackage{tabularx}
\usepackage{array}
\usepackage{setspace}

\newcolumntype{M}[1]{>{\centering\arraybackslash}m{#1}}
\newcommand{\relposiset}{\tilde{\mathbf{X}}}
\newcommand{\relposi}{\tilde{\mathbf{x}}}
\newcommand{\relposimirror}{\check{\mathbf{x}}}
\newcommand{\anchorposi}{\tilde{\mathbf{x}}}
\newcommand{\anchorposiset}{\tilde{\mathbf{X}}}
\newcommand{\globalposi}{\mathbf{x}}

\newcommand{\errorvec}{\mathbf{e}}

\usepackage{caption}
\usepackage{subcaption}

\def\BigRoman{\uppercase\expandafter{\romannumeral\number\count 255 }}
\def\Romannumeral{\afterassignment\BigRoman\count255=}
\def\figref#1{Fig.~\ref{#1}}

\def\eqref#1{(\ref{#1})}

\def\argmin{\mathop{\rm argmin}}
\definecolor{lightgreen}{rgb}{0.25 ,0.5, 0.25}
\definecolor{skyblue}{rgb}{0.25, 0.25, 0.8}

\title{\LARGE \bf
SaWa-ML: Structure-Aware Pose Correction and Weight Adaptation-Based Robust Multi-Robot Localization}

\author{Junho Choi$^{1}$, Kihwan Ryoo$^{1}$, Jeewon Kim$^{1}$, Taeyun Kim$^{1}$, Eungchang Lee$^{1}$, Myeongwoo Jeong$^{1}$, \\ Kevin Christiansen Marsim$^{1}$, Hyungtae Lim$^{2}$, and Hyun Myung$^{1*}$, \textit{Senior Member, IEEE}
\thanks{$^*$Corresponding author: Hyun Myung}
\thanks{$^{1}$Junho Choi, Kihwan Ryoo, Jeewon Kim, Taeyun Kim, Eungchang Lee, Myeongwoo Jeong, Kevin Christiansen Marsim, Hyun Myung are with the School of Electrical Engineering, KAIST (Korea Advanced Institute of Science and Technology), Daejeon, 34141, Republic of Korea. Email: {\tt\scriptsize \{cjh6685kr, rkh137, ddarong2000, ktw1404, eungchang\_mason, wjdauddn1477, kevinmarsim, hmyung\}@kaist.ac.kr} \hfill \break 
\indent $^{2}$Hyungtae Lim is with the Laboratory for Information \& Decision Systems (LIDS), Massachusetts Institute of Technology, Cambridge, MA02139, USA, Email: {\tt\scriptsize \{shapelim\}@mit.edu} \hfill \break
\indent This work has been supported by the Unmanned Swarm CPS Research Laboratory Program of Defense Acquisition Program Administration and Agency for Defense Development(UD220005VD). The students are supported by BK21 FOUR.}
}

\begin{document}
\maketitle
\thispagestyle{empty}
\pagestyle{empty}

\begin{abstract}
Multi-robot localization is a crucial task for implementing multi-robot systems. Numerous researchers have proposed optimization-based multi-robot localization methods that use camera, IMU, and UWB sensors. Nevertheless, characteristics of individual robot odometry estimates and distance measurements between robots used in the optimization are not sufficiently considered. In addition, previous researches were heavily influenced by the odometry accuracy that is estimated from individual robots. Consequently, long-term drift error caused by error accumulation is potentially inevitable. In this paper, we propose a novel visual-inertial-range-based multi-robot localization method, named \textit{SaWa-ML}, which enables geometric structure-aware pose correction and weight adaptation-based robust multi-robot localization. Our contributions are twofold: (i)~we leverage UWB sensor data, whose range error does not accumulate over time, to first estimate the relative positions between robots and then correct the positions of each robot, thus reducing long-term drift errors, (ii)~we design adaptive weights for robot pose correction by considering the characteristics of the sensor data and visual-inertial odometry estimates. The proposed method has been validated in real-world experiments, showing a substantial performance increase compared with state-of-the-art algorithms.

\end{abstract}

\section{Introduction}
As demand for robotic applications increases, the deployment of multi-robot systems for complex tasks becomes more essential. Extensive research has been conducted to optimize multi-robot systems, which are more versatile than single-robot systems for complex tasks. Particularly, accurate localization of robots is essential for the effective utilization of multi-robot systems. Transitioning from single-robot to multi-robot systems necessitates accurate estimation of the relative poses between robots and their individual poses within a unified world frame.
\begin{figure}[t]
\centering  
{\includegraphics[width=0.99\linewidth]{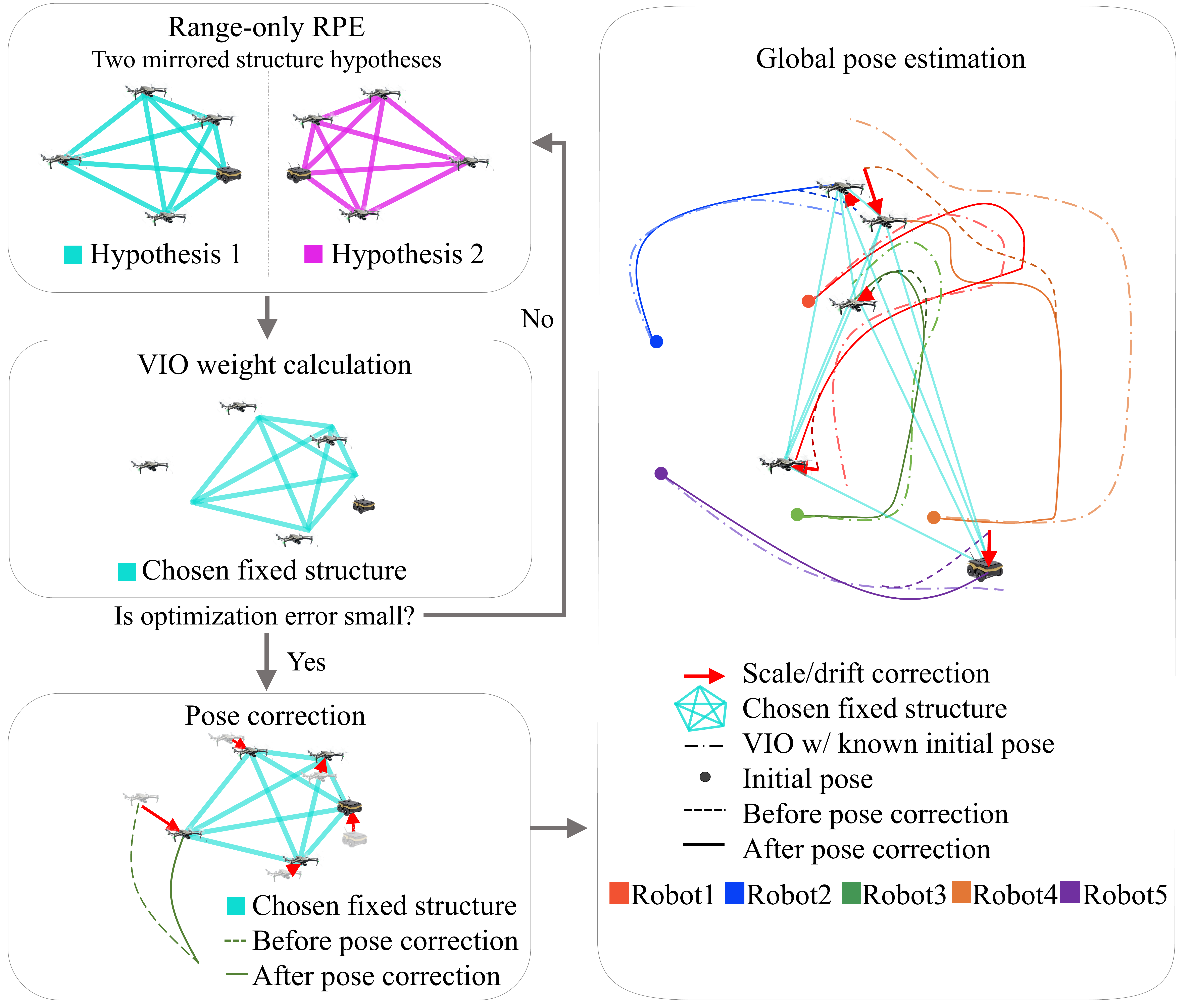}}
\captionsetup{font=footnotesize}
\caption{This figure describes the flow of the proposed method with five robots. The geometric structure of the relative position of the robots is fixed with UWB range measurements to avoid long-term drift of VIO. Using the range measurements, two structures can be made that are mirrored to each other (Section \Romannumeral3.C). Then, the two structures are transformed into the world frame using adaptive weight optimization and one of them is selected (Section \Romannumeral3.D). The drift and scale error of the VIO are estimated and the pose nodes are corrected (Section \Romannumeral3.E). The global poses of the multi-robot are estimated at the same rate as the local VIO computation period (Section \Romannumeral3.B).}
\label{first_figure}
\end{figure}

Although global navigation satellite system (GNSS)-based positioning methods are prevalent and straightforward to implement, they are ineffective in indoor environments and are susceptible to jamming and spoofing attacks. Consequently, research into non-GNSS-based robot localization has become essential. Methods that use motion tracking systems\cite{preiss2017crazyswarm}, or pre-installed ultra-wideband (UWB) anchors\cite{single_UAV_odom1, single_UAV_odom2} are capable of accurately estimating robot poses without using GNSS. However, these methods are generally limited by their operation area and the need for additional infrastructure, which are significant drawbacks for widespread deployment. Thus, it is crucial to employ exteroceptive sensors such as cameras and LiDARs, alongside proprioceptive sensors like encoders and inertial measurement units (IMUs), mounted on the robots to estimate their poses accurately. As the scale of multi-robot systems expands with the inclusion of more robots, visual-inertial-based methods are preferred over LiDAR-based methods for localization due to the high cost of LiDARs. Multi-robot localization methods based on visual-inertial sensors are proposed in \cite{single_UAV_odom1, single_UAV_odom2}. These methods share visual features among the robots, using them to estimate relative poses through loop detection and pose correction. However, visual-inertial-based localization methods may not provide sufficient accuracy, particularly in scenarios where robots traverse long trajectories, due to long-term drift or scale errors. To address the limitations of visual-inertial-based localization methods, additional range sensors are used widely. Various visual-inertial-range-based multi-robot localization methods\cite{multi-UAV_local_3_nguyen2021flexible, multi-UAV_local_15_9896952, multi-UAV_local_14_10043709, multi-UAV_local_6_queralta2020vio} have been developed to expand upon visual-inertial-range localization methods for a single robot \cite{UVIO1_nguyen2020tightly, UVIO2_nguyen2021range, UVIO3_yang2021uvip, UVIO4_jung2022u, UVIO6_delaune2021range, UVIO7_yang2021resilient, UVIO8_wang2017ultra, UVIO9_lutz2019visual, UVIO10_10301588}. These methods utilize additional measurements of distances among robots to reduce scale and drift errors. Despite the variety of proposed solutions, there remains a need for a multi-robot localization method that is robust even when some robots experience significant long-term drift or scale errors.

In this paper, we introduce \textit{SaWa-ML} (structure-aware pose correction and weight adaptation-based robust multi-robot localization), a novel method specifically designed to increase the accuracy of multi-robot localization. This method proves particularly effective in scenarios where previous methods are challenged by inaccuracies that accumulate over extended trajectories. The overall framework of the proposed method is illustrated in \figref{first_figure}. The robots continuously exchange data, including UWB range measurements and visual-inertial odometry (VIO). Then, through range-only relative position estimation, the geometric structure of the multi-robot is estimated. In addition, adaptive weight calculation considering VIO characteristics and global consistency of shared measurements, the pose of each robot is robustly estimated. The main contributions of this paper are as follows:
\begin{itemize}
    \item {We propose a method that utilizes UWB range data, whose error is not accumulated over time, to calculate the geometric structure formed by the positions of the robots and then use this structure to correct their positions, thereby reducing long-term drift error.}
    \item {We adjust weights adaptively during the correction phase, considering the characteristics of VIO and UWB range data as well as global consistency of the measurements and estimates.}
    \item {The proposed methods is validated with real-world experiments on longer trajectories compared with those tested in state-of-the-art researches.}
\end{itemize}

The rest of the paper is organized as follows. In Section {\Romannumeral 2}, related works on multi-robot localization methods based on visual-inertial range are reviewed. Then, the proposed method is explained in detail in Section {\Romannumeral 3}. In Section {\Romannumeral 4}, experimental environments and results are introduced. Lastly, we conclude in Section {\Romannumeral 5}.

\section{Related Works}
Various localization methods for multi-robot systems have been proposed\cite{multi-UAV_local_1_xu2022omni, multi-UAV_local_2_li2018accurate, multi-UAV_local_3_nguyen2021flexible, multi-UAV_local_4_ruan2021cooperative, multi-UAV_local_5_chakraborty2019cooperative, multi-UAV_local_6_queralta2020vio, multi-UAV_local_7_papadimitriou2022range, multi-UAV_local_8_nguyen2018robust, multi-UAV_local_9_guo2019ultra, multi-UAV_local_11_cvi, multi-UAV_local_12_covins, multi-UAV_local_13_dubois2019data, multi-UAV_local_14_10043709, multi-UAV_local_15_9896952, UVIO5_sivaneri2017ugv}. The method using fixed UWB anchors \cite{multi-UAV_local_2_li2018accurate} shows accurate pose estimation by utilizing known positions of pre-installed anchors, which can be only used within the specific bounded areas. Hence, it loses the advantage of simultaneous mission performance over a wide area that a multi-robot system has.

To overcome the drawback, multi-robot localization methods that share visual information without using pre-installed UWB anchors have also been proposed\cite{multi-UAV_local_12_covins, multi-UAV_local_11_cvi, visualmarker1, visualmarker2, visualmarker3}. In \cite{multi-UAV_local_11_cvi} and \cite{multi-UAV_local_12_covins}, keyframes or visual feature data extracted from the images are shared among the robots and used to estimate relative poses between the robots. However, vision-based methods are very vulnerable to domain differences in captured images from each robot. Even if the images are taken from the same location, visual information matching among the robots may fail owing to image differences caused by differences of camera properties, viewpoint, and illuminance. Additionally, vision-based loop closing is available only when the robots visit the same location or see the same view. Therefore, each robot has to inevitably rely on its own local VIO that is calculated by itself until loop closing occurs. Because each VIO has long-term drift and scale error, relying only on visual information is not sufficient for multi-robot localization. Robot detection-based multi-robot localization methods\cite{visualmarker1,visualmarker2, visualmarker3} also have similar disadvantages that relative pose estimation is possible only when the robot is captured by another robot's camera.

Additional researches have been conducted using UWB sensors, which are not susceptible to long-term drift over time\cite{multi-UAV_local_3_nguyen2021flexible, multi-UAV_local_4_ruan2021cooperative, multi-UAV_local_14_10043709, multi-UAV_local_6_queralta2020vio, multi-UAV_local_15_9896952}. Unlike the method that uses UWB anchors \cite{multi-UAV_local_2_li2018accurate}, visual-inertial-range-based multi-robot localization methods \cite{multi-UAV_local_3_nguyen2021flexible, multi-UAV_local_4_ruan2021cooperative, multi-UAV_local_14_10043709, multi-UAV_local_6_queralta2020vio, multi-UAV_local_15_9896952} are not limited by the operating area, because the UWB sensor is attached to each robot and used solely to measure the distance among the robots. The methods introduced in \cite{multi-UAV_local_3_nguyen2021flexible,multi-UAV_local_15_9896952, multi-UAV_local_14_10043709} showed more accurate localization results compared with individual VIO estimates. However, the experimental results did not sufficiently demonstrate pose correction, because the operation time was short with the short trajectories and only accurate VIO data were used. The VIO method must be considered for its vulnerability to long-term drift and scale error, and it is necessary to verify whether the proposed multi-robot localization methods can be operated robustly even when drift or scale errors occur in some of the robots.

In summary, despite the proposal of various multi-robot localization methods, they still require fixed infrastructure, are sensitive to environmental factors in vision-based systems, or the multi-robot localization results are heavily influenced by individual robot's localization estimates, presenting ongoing limitations. Therefore, we aim to develop an accurate range-aided visual-inertial multi-robot pose estimation method that leverages the characteristics of multi-robot systems.

\begin{figure*}[t]
\centering  
{\includegraphics[width=1.0\linewidth]{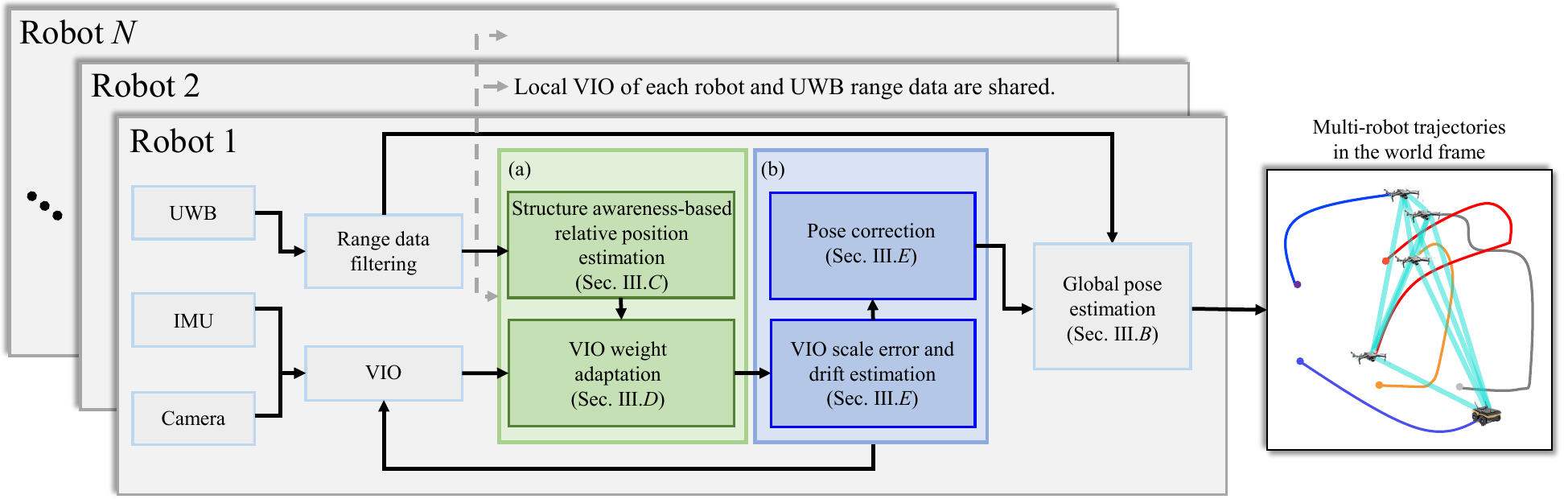}}
\captionsetup{font=footnotesize}
\caption{Overview of the proposed multi-robot localization framework, called \textit{SaWa-ML}. VIO is calculated in each robot using an attached IMU and camera. Filtered range data and the VIO calculated in each robot are shared among all robots. In the (a) structure awareness and weight adaptation stage, the relative positions of the robots, called structure, and weights are calculated. Then, pose error of each robot is calculated and corrected in the (b) error estimation and pose correction stage. Subsequently, global pose estimation is performed using the corrected VIO and the filtered range data.
}
\label{overview}
\end{figure*}

\section{Structure awareness and weight adaptation-based multi-robot localization}
In contrast to single robot localization, multi-robot localization employing a visual-inertial-range sensor system leverages additional data, such as each robot's local odometry and inter-robot distance measurements. Specifically, local VIO and UWB measurement exhibit complementary properties: while UWB sensors are susceptible to non-line-of-sight (NLOS) conditions but have the advantage that the range error does not accumulate over time; Conversely, local VIO from each robot remains unaffected by NLOS yet suffers from odometry error accumulation over time, which is known as long-term drift, and potential scale discrepancies arising from camera modeling uncertainty. Our proposed method integrates VIO and UWB range measurements, aiming to robustly and accurately estimate multi-robot poses by maximizing the strengths and minimizing the weaknesses of each data.

\subsection{System Overview}

The overall framework of the proposed method is shown in \figref{overview}. Each robot is equipped with a UWB sensor that can be used as both an anchor and a tag, an IMU, and a stereo camera. The UWB sensors measure the distances between each pair of robots. The measured distances are initially smoothed with a moving average filter to mitigate noises, followed by a random sample consensus (RANSAC) process to exclude outliers. Local odometry of each robot is estimated by a VIO method with the measurements of the IMU and the stereo camera mounted on the robots. Among many proposed VIO methods, we employed VINS-Fusion \cite{single_UAV_odom1}, which was also used in one of the comparative methods. However, any other VIO methods can be used without loss of generality.

In the structure awareness and weight adaption step, the local VIO and the filtered UWB data are shared among robots. Then, the geometric structure (relative positions between robots) is estimated using solely the filtered UWB data, preventing the accumulation of VIO errors. This estimated structure is then utilized for error estimation and pose correction if it satisfies predefined conditions. Global consistency of the shared data is evaluated by comparing actual range measurements with the predicted distances that are calculated from the estimated distances from VIO estimates. Due to the vulnerability of the VIO to rapid movements, weights for the correction step are calculated considering the previously computed global consistency along with each robot's linear velocity and angular velocity.

The scale error and drift, which are critical factors affecting the accuracy of the local VIO, are calculated from the global poses. These errors are then used to correct the pose of each robot. Furthermore, the calculated scale error and drift are fed back into the VIO estimation step of each robot. Then, the global poses of all robots are optimized using the corrected VIO on the world frame, and shared range data. 


\subsection{Global Pose Estimation}
The optimization for the global pose estimation runs with the same interval of the local VIO computation interval $\Delta t$. The local VIO that is estimated at the local frame of each robot and UWB range measurements are shared, then the poses of the $N$ robots in the world frame ${^{W}}\textbf{T}_{t}=\{{^{W}}\textbf{T}^{(1)}_{t}, \cdots, {^{W}}\textbf{T}^{(N)}_{t}\}$ where ${^{W}}\textbf{T}^{(i)}_{t}\in$ SE(3), are estimated by optimizing the equation as follows:
\begin{multline}
        	\argmin\limits_{{^{W}}{\globalposi}_{t}^{(i)}, i=1, 2, \cdots, N }\sum_{i=1}^{N}\bigg\{ \sum_{j=1}^{N}|r_{ij;t}^{2}-||{^{W}}{\globalposi}^{(i)}_{t}-{^{W}}{\globalposi}^{(j)}_{t}||^{2}|\\
         + ||{{^{W}}\textbf{T}^{(i) -1}_{0}}{{^{W}}\textbf{T}^{(i)}_{t}} -{^{L}\textbf{T}^{(i)}_{t}}||_{F}^{2}\bigg\},
\end{multline}
where $r_{ij;t}$ is the filtered range data between Robot $i$ and Robot $j$ at time $t$, and ${^{W}}{\globalposi}_{t}^{(i)}=({^{W}}x_{t}^{(i)}, {^{W}}y_{t}^{(i)}, {^{W}}z_{t}^{(i)})$ denotes the estimated position of Robot $i$ in the world frame $W$ at time $t$. Absolute value, $l^2$ norm, and Frobenius norm are denoted by $\left| \cdot \right|$, $\left\| \cdot \right\|$ and $\left\| \cdot \right\|_{F}$, respectively. The world frame $W$ is set to the local frame of Robot 1, so the origin of the world frame is $^{L}{\globalposi}^{(1)}_{0}$, where the local VIO frame of each robot is denoted by $L$ and $^{L}{\globalposi}^{(i)}_{0} = (0, 0, 0)$.

\subsection{Range-Only Multi-Robot Structure Awareness}
\begin{figure*}[t]
\centering  
{\includegraphics[width=1\linewidth]{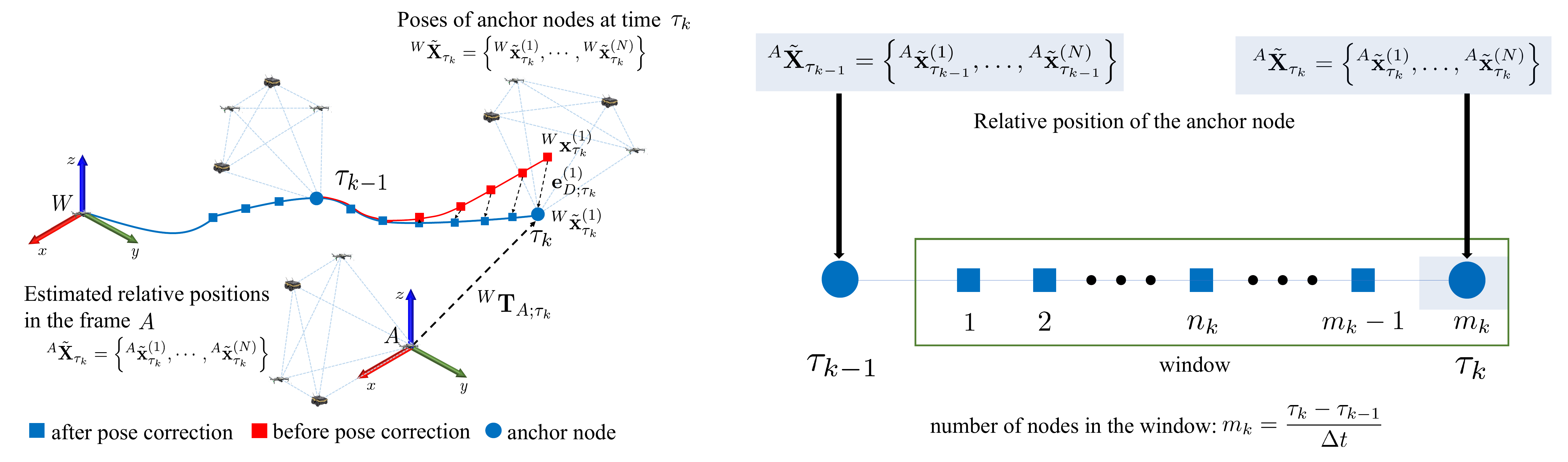}}
\captionsetup{font=footnotesize}
\caption{An example of the proposed pose correction process at $\tau_k$. The relative positions of the robots are estimated and fixed at the range-only relative position estimation step. Then the fixed structure is transformed into the world frame (cyan dotted line) using calculated weights and the pose of each robot becomes an anchor node (blue circles). When an additional anchor node is generated, the global pose estimation nodes (red squares) between the previous anchor node and the current anchor node are corrected (blue squares).}
\label{pose_correction}
\end{figure*}

One of the important characteristics of UWB range data is that there is no error accumulation over time and no drift occurs owing to the movement of robots. Therefore, it can be effective to determine the relative position-based geometric structure only with UWB range data without using the local VIO which contains accumulated error. The goal of this step is not to estimate the relative pose in the world frame of the multi-robot system, but to estimate the relative positions of the robots in frame $A$, which is introduced for relative position estimation. The frame $A$ is generated by fixing Robot 1 at the origin, Robot 2 on the $x$-axis, and Robot 3 on the $xy$-plane. When each robot position is set as a vertex and the range measurements are set as a linkage with a fixed length connecting each vertex, the geometric structure generated by these vertices and linkages is fixed. For $N$ robots, $\frac{N(N-1)}{2}$ range data are obtained. The more robots are used, the more geometrical constraints can be generated. The relative position structure of the robots is estimated by optimizing the residual as follows:
\begin{equation}
\label{rel posi eq}
    	\ \argmin \limits_{^{A}{\relposi}^{i}_{t}, i=1, 2, \cdots, N}{ \sum_{i=1}^{N}\sum_{j=1(i \neq j)}^{N}|r_{ij;t}^{2}-||^{A}{\relposi}^{(i)}_{t}-{^{A}}{\relposi}^{(j)}_{t}||^{2}|},
\end{equation}
where ${^{A}}{\relposi}_{t}^{(i)}$ denotes the position (${^{A}}x_{t}^{(i)}, {^{A}}y_{t}^{(i)}, {^{A}}z_{t}^{(i)}$) of Robot $i$ at time $t$ in the coordinate frame $A$. In the optimization of \eqref{rel posi eq}, the result of calculating the defined residual summand is denoted as $\boldsymbol{r}_{r;t}$. The optimized estimates are only used when the residual $\boldsymbol{r}_{r;t}$ of optimization is less than the user-defined threshold $\zeta$. This is because UWB sensor measurement is noisy and large error may occur in NLOS situations. In a multi-robot system, even if there are no NLOS-causing obstacles in the environment, NLOS situations may occur owing to their own body frames of the robots. When the criterion is met, the range-only relative position set is used in the adaptive VIO weight calculation step. The $k$-th relative positions of the robots in the frame $A$ are denoted by ${^{A}}{\relposiset}_{\tau_{k}}=\{{^{A}}{\relposi}^{(1)}_{\tau_{k}}, \cdots, {^{A}}{\relposi}^{(N)}_{\tau_{k}}\}$ as shown in \figref{pose_correction}, where $\tau_{k}$ is the time when $k$-th set was generated. The positions of robots that were estimated using only ranges have ambiguity about the mirror structure that is denoted by $^{A}\relposimirror_{\tau_{k}}$.

\subsection{Adaptive VIO Weight Calculation}
When $^{A}{\relposiset}=\{^{A}{\relposi}^{(1)}, \cdots, {^{A}}{\relposi}^{(N)}\}$ meets the criteria mentioned in the previous subsection, the transformation matrix $^{W}\textbf{T}_{A}$ is calculated. This matrix transforms the position matrix of the frame $A$ to the world frame $W$ using adaptive VIO weight calculation as follows:
\begin{equation}
    \label{eq4}
    \underset{{^{W}}\textbf{T}_{A; \tau_{k}}}{\argmin}\sum_{i=1}^{N}{w_{\tau_{k}}^{(i)}}\left\| {^{W}}{\globalposi}_{\tau _{k}}^{(i)} - {^{W}}\textbf{T}_{A; \tau_{k}}{^{A}}{\relposi}_{\tau _{k}}^{(i)}\right\|.
\end{equation}
The weight $w$ is designed considering the characteristics of the VIO and the global consistency of the measurements. As mentioned in \cite{multi-UAV_local_3_nguyen2021flexible, drift}, the drift of VIO occurs largely in the direction in which the robot moves. Additionally, VIO is vulnerable to loss of visual feature tracking. We observed that the visual-inertial odometry is more likely to drift with rotational motions not only in our data but also from the VIO benchmark paper \cite{single_dataset4}. The comparison results of \cite{single_dataset4} show that even in the same path, the more the robot rotates, the larger the error becomes. Thus, we designed velocity weight $w_v$ and rotational motion weight $w_a$ taking into account the visual features that can be lost as the linear velocity $v$ or angular velocity $\omega$ increases as follows:
\begin{equation}
\label{eq5}
    \scalebox{0.87}{
    $
    \begin{aligned}
    w^{(i)}_{v ;\tau _{k}}=\frac{1}{m_{k}}\sum_{i=1}^{m_{k}}&\left\{ \left ( \frac{\textrm{max}(l_{\tau _{k-1}+i\Delta t}-\textrm{max}({v}_{x;\tau _{k-1}+i\Delta t}, 0)\Delta t, 0)}{l_{\tau _{k-1}+i\Delta t}} \right )^{2} \right. \\
    &\;\;\, \left. + \frac{\textrm{max}(l_{\tau _{k-1}+i\Delta t} - \left| {v}_{y;\tau _{k-1}+i\Delta t}\right|\Delta t, 0)}{l_{\tau _{k-1}+i\Delta t}} \, \right.\\
    &\;\;\, \left. + \frac{\textrm{max}(l_{\tau _{k-1}+i\Delta t} - \left| {v}_{z;\tau _{k-1}+i\Delta t}\right|\Delta t, 0)}{l_{\tau _{k-1}+i\Delta t}}\right\}, 
    \end{aligned}
    $
    }
\end{equation}
where $l$ is the mean depth of visual features back-projected into a 3D space,
\begin{align}
w^{(i)}_{a;\tau_k} &= \frac{1}{m_k} \sum_{i=1}^{m_k} \left( \frac{\max(\theta_h - \omega^{(i)}_{h,\tau_{k-1} + i\Delta t} \Delta t, 0)}{\theta_h} \right. \nonumber \\
&\qquad\quad\qquad\; \left. \cdot \frac{\max(\theta_v - \omega^{(i)}_{v,\tau_{k-1} + i\Delta t} \Delta t, 0)}{\theta_v} \right),
\end{align}
where $\theta_h$ and $\theta_v$ are the horizontal and vertical field of views of the camera, respectively. The number of pose node from $\tau _{k-1}$ to $\tau _{k}$ is denoted by $m_{k}$ and calculated as follows:
\begin{equation}
    \label{eq9}
    m_{k}=\frac{\tau _{k} - \tau _{k-1}}{\Delta t}.
\end{equation}
Our idea for maintaining global consistency on every VIO in the world frame comes from the observation that the accurately estimated VIO gives a distance similar to that from range-only relative position estimation. For instance, if the VIO of Robot 1 and the VIO of Robot 2 are accurate enough, the distance between positions of the robots that are estimated by VIO in the world frame should be similar to the distance between them on the range-only relative position estimation. Thus, we designed global consistency weight $w_r$ as follows:
\begin{equation}
    \label{eq7}
    w^{(i)}_{r;\tau _{k}}=\frac{1}{m_{k}}\sum_{h=1}^{m_{k}}\sum_{i=1}^{N}\sum_{\underset{(i\neq j)}{j=1}}^{N}\frac{\tilde{d}_{ij;{\tau _{k-1}+h\Delta t}}}{\left| d_{ij;{\tau _{k-1}+h\Delta t}}-\tilde{d}_{ij;{\tau _{k-1}+h\Delta t}} \right|+\epsilon},
\end{equation}
where $d_{ij;\tau _{k}} = \left\|{^{W}}{\globalposi}^{(i)}_{\tau_{k}}-{^{W}}{\globalposi}^{(j)}_{\tau_{k}}\right\|$, $\tilde{d}_{ij;\tau _{k}} = \left\|{^{A}}{\relposi}^{(i)}_{\tau_{k}}-{^{A}}{\relposi}^{(j)}_{\tau_{k}}\right\|$ and $\epsilon$ is a positive infinitesimal value to prevent division by zero.

The calculated three types of weights ${w}_{v}$, ${w}_{a}$ and ${w}_{r}$ are normalized by dividing each value by the maximum value among them, ensuring that the maximum normalized value becomes 1. These normalized weights are denoted as $\hat{w}_{v}$, $\hat{w}_{a}$ and $\hat{w}_{r}$, respectively.
Then, the overall weight of Robot $i$ at time $\tau_k$ is the sum of the three weights as follows:
\begin{equation}
    \label{eq8}
    w_{\tau_{k}}^{(i)} = \hat{w}_{a;\tau_{k}}^{(i)}+\hat{w}_{v ;\tau_{k}}^{(i)}+\hat{w}_{r;\tau_{k}}^{(i)}.
\end{equation}
The same calculation is also carried out for the mirror structure $^{A}{\relposimirror}_{\tau_{k}}$, and one of the two mirror structures having less error is determined as $^{A}{{\relposi}}_{\tau_{k}}$. As a result, the $k$-th anchor node of Robot $i$ at ${^{W}}{\anchorposi}^{(i)}_{\tau_{k}}=({^{W}}\tilde{x}_{\tau_k}^{(i)}, {^{W}}\tilde{y}_{\tau_k}^{(i)}, {^{W}}\tilde{z}_{\tau_k}^{(i)})$, as described in \figref{pose_correction}, is calculated as follows:
\begin{equation}
{^{W}}{\anchorposi}^{(i)}_{\tau_{k}}={^{W}}\textbf{T}_{A; \tau_{k}}{^{A}}{\relposi}_{\tau _{k}}^{(i)}.    
\end{equation}
When the $k$-th anchor nodes ${^{W}}{\anchorposiset}_{\tau_{k}} = \{{^{W}}{\anchorposi}^{(1)}_{\tau_{k}}, \cdots, {^{W}}{\anchorposi}^{(N)}_{\tau_{k}}\}$ are generated at $\tau_k$, the $m_{k}$ poses from $\tau_{k-1}$ to $\tau_k$ that were estimated in the global pose estimation step are corrected as in the following subsection.

\subsection{Pose Correction Using Anchor Nodes}
From $\tau_2$, which is the time when the second anchor nodes are generated, the estimated nodes ${^{W}}{\globalposi}$ between two anchor nodes of all the robots are corrected at every $\tau_k$ as shown in \figref{pose_correction}. When the anchor nodes ${^{W}}{\anchorposi}_{\tau_k}$ are calculated, scale error ${\errorvec}_{S_{\tau_k}}^{(i)} = (e^{(i)}_{S_{x;\tau_k}}; e^{(i)}_{S_{y;\tau_k}}; e^{(i)}_{S_{z;\tau_k}})$ of ${^{W}}{{\globalposi}}^{(i)}_{\tau_k}$ from $\tau_{k-1}$ to $\tau_k$ is calculated using previous anchor node and current anchor node. Each direction has a different scale error, thus the scale error ${\errorvec}_{S}$ of Robot $i$ at time $\tau_k$ is calculated as follows:
\begin{equation}
    \label{eq11}
    {\errorvec}_{S;\tau_k}^{(i)}=\begin{pmatrix}
e^{(i)}_{S_{x};\tau_k} \\[8pt]
e^{(i)}_{S_{y};\tau_k} \\[8pt]
e^{(i)}_{S_{z};\tau_k}
\end{pmatrix}
=\begin{pmatrix}
\frac{| {^{W}}\tilde{x}^{(i)}_{\tau_k} - {^{W}}\Tilde{x}^{(i)}_{\tau_{k-1}}|}{| {^{W}}x^{(i)}_{\tau_k} - {^{W}}x^{(i)}_{\tau_{k-1}}|}\\[8pt]
\frac{| {^{W}}\tilde{y}^{(i)}_{\tau_k} - {^{W}}\Tilde{y}^{(i)}_{\tau_{k-1}}|}{|{^{W}}y^{(i)}_{\tau_k} - {^{W}}y^{(i)}_{\tau_{k-1}}|}\\[8pt]
\frac{| {^{W}}\tilde{z}^{(i)}_{\tau_k} - {^{W}}\Tilde{z}^{(i)}_{\tau_{k-1}}|}{|{^{W}}z^{(i)}_{\tau_k} - {^{W}}z^{(i)}_{\tau_{k-1}}|}
\end{pmatrix}.
\end{equation}
At $\tau_k$, ${{^{W}}{{\globalposi}}^{(i)}_{\tau_{k-1}}}=({^{W}}{x}_{\tau_k-1}^{(i)}, {^{W}}{y}_{\tau_k-1}^{(i)}, {^{W}}{z}_{\tau_k-1}^{(i)})$ and ${^{W}}{\anchorposi}^{(i)}_{\tau_{k-1}}=({^{W}}\tilde{x}_{\tau_k-1}^{(i)}, {^{W}}\tilde{y}_{\tau_k-1}^{(i)}, {^{W}}\tilde{z}_{\tau_k-1}^{(i)})$ are identical owing to the pose correction. All the values of the scale error factors are initialized to 1 so that ${\errorvec}_{S;0}^{(i)}=(1, 1, 1)^T$. Before pose correction, position difference ${\errorvec}_{D;\tau_k}$ exists between ${^{W}}{\anchorposi}$ and ${^{W}}{\globalposi}$ every time when ${^{W}}{\anchorposi}$ is determined. Then, the poses from $\tau_{k-1}$ to $\tau_k$ are corrected as follows:
\begin{equation}
        {^{W}}{\anchorposi}_{\tau_{k-1}+n_{k}\Delta t}={^{W}}{\globalposi}_{\tau_{k-1}+n_{k}\Delta t} + \frac{n_{k}}{m_{k}}{\errorvec}_{D;\tau_k},
\end{equation}
where ${\errorvec}_{D;\tau_k} = {^{W}}{\anchorposi}_{\tau_k} - {{^{W}}{\globalposi}_{\tau_k}}$
and $0<n_{k}<m_{k}$.
Then, the scale error factor is updated to ${\errorvec}_{S;\tau_{k-1}}^{(i)} \odot {\errorvec}_{S;\tau_k}^{(i)}$ where $\odot$ denotes Hadamard product. The scale error factor ${\errorvec}_{S}$ is fedback to each robot for local VIO correction. From $\tau_{k}+\Delta t$ to $\tau_{k+1}-\Delta t$, the VIOs of the robots are multiplied by the scale error ${\errorvec}_{S;\tau_k}^{(i)}$.

\section{Experiments}
\subsection{Experimental Environments}
\begin{figure}[t]
\centering  
\subfigure[]{\includegraphics[width=0.99\linewidth]{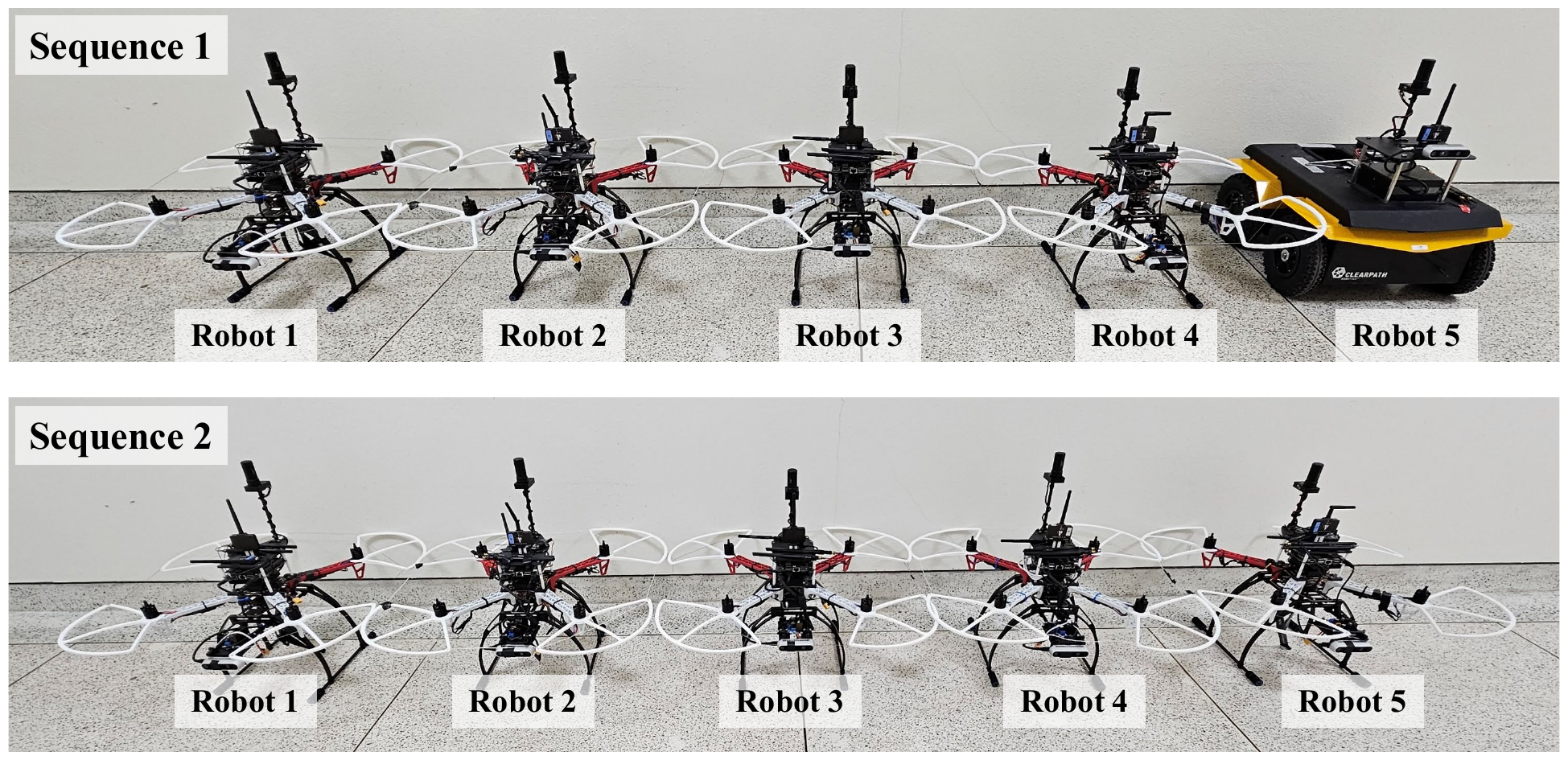}}
\subfigure[]{\includegraphics[width=0.32\linewidth]{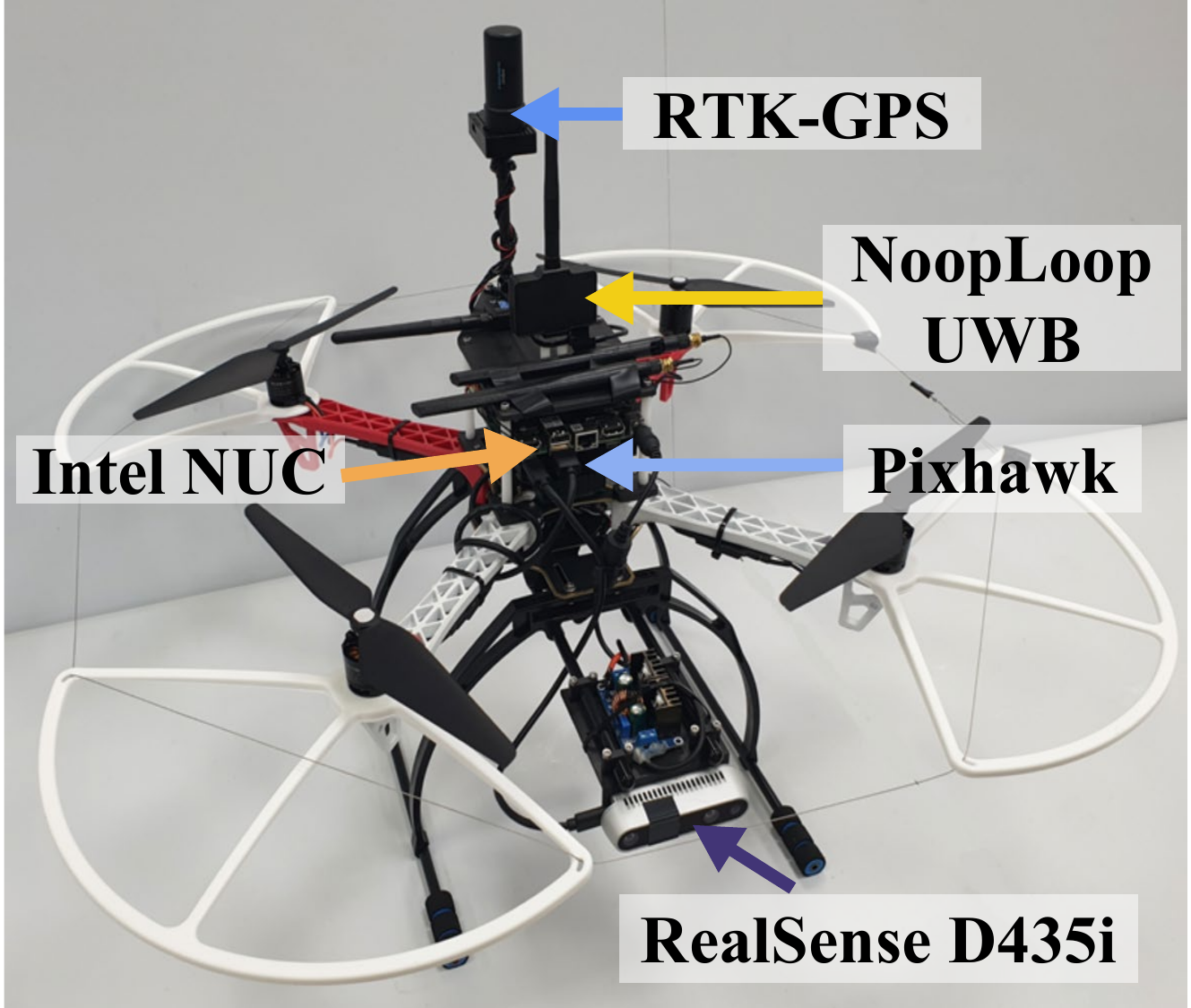}}
\subfigure[]{\includegraphics[width=0.32\linewidth]{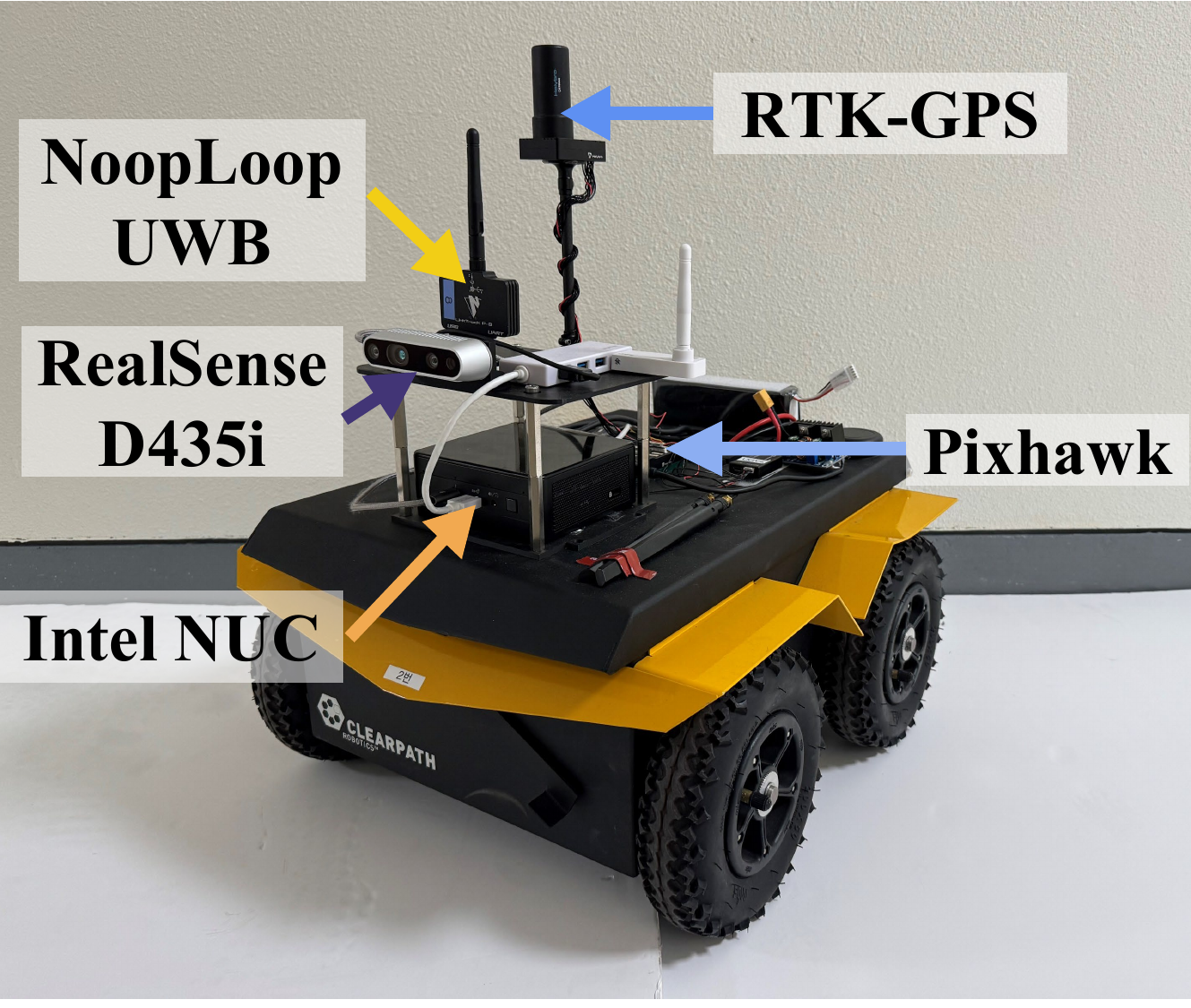}}
\subfigure[]{\includegraphics[width=0.32\linewidth]{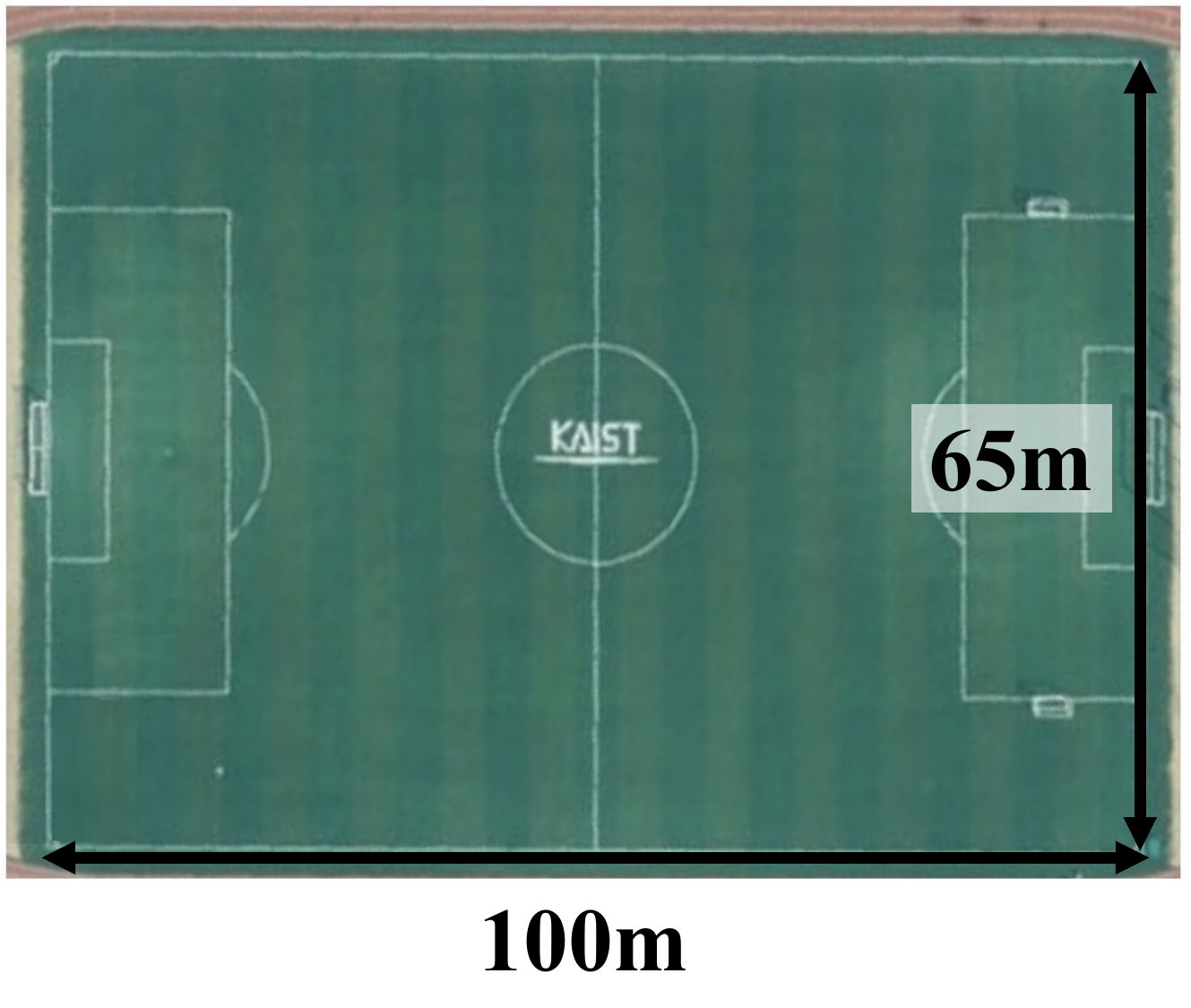}}
\captionsetup{font=footnotesize}
\caption{The customized UAVs and UGVs for the experiments. (a) Multi-robot systems that were used to get Sequence~1 and Sequence~2. (b) A sensor configuration of the UAV. Pixhawk6c is attached to the center of the UAV and Intel NUC is located on it. The Nooploop LinkTrack P-B UWB is mounted on the UAV. The Intel Realsense camera D435i is attached at the front of the UAV. (c) A sensor configuration of the UGV. Sensors with the same configuration as the UAV are also mounted on the UGV. (d) All experiments were conducted in a large outdoor environment shown here (Top view).}
\label{UAV_sensor_setup}
\end{figure}

Unfortunately, multi-robot datasets that contain all of visual, inertial, and range (among the robots), and ground-truth information do not exist at the moment, hence existing state-of-the-art methods were also tested with their own data. Therefore, we obtained corresponding data using the customized multi-robot system as shown in \figref{UAV_sensor_setup}(a).
Four UAVs and one UGV were used in Sequence~1 and five UAVs were used in Sequence~2. More details of the UAV and UGV platforms are shown in Figs.~4(b) and (c), respectively. Both of the robots are equipped with a stereo camera, an IMU, and a UWB. The real-time kinematic GPS (RTK-GPS) is attached to each robot for ground truth measurement. H-RTK F9P Helical with centimeter-accuracy is used. Range data are logged at a rate of 50 Hz using Nooploop LinkTrack P-B UWB mounted on the UAVs. The UWB module acts as both an anchor and a tag, so each UAV obtains the range data of all other UAVs. The sensor's time synchronization is provided from the UWB. As well explained in \cite{multi-UAV_local_1_xu2022omni}, the UWB modules provide synchronized timestamp and the UWB module is directly connected to the onboard computer through the serial port. It synchronizes the time stamps during the communications. Stereo image data with 640 $\times$ 480 resolution at 30 Hz are obtained by Intel Realsense D435i facing directly forward. The stereo images are used to calculate VIO. An IMU is mounted at the center of the robot and IMU data are logged at a rate of 100 Hz using Pixhawk6c. The intrinsic matrix of the camera and the extrinsic matrix between the camera and IMU were calculated with Kalibr\cite{kalibr}. Real-world experiments were conducted in a large outdoor environment of 100 m $\times$ 65 m as shown in \figref{UAV_sensor_setup}(d).

\subsection{Experimental Results}
Many visual-inertial-range-based multi-robot localization methods have been proposed. However, the source codes and experimental data used for performance evaluation have not been made publicly available at the moment. Therefore, the state-of-the-art methods were implemented from the papers as similar as possible and used for comparison. The evaluation was conducted using two sequences obtained from outdoor experiments. Quantitative values were compared with the RMSE of absolute trajectory error that was calculated using EVO\cite{grupp2017evo}. Multi-robot localization methods aim to estimate the pose of multiple robots in the same world frame. Therefore, the calculation of all absolute trajectory errors was performed by only overlapping each estimated initial pose with its corresponding ground truth. A comparison was conducted in two scenarios: (1) without known initial pose and (2) with known initial pose with two sequences.

\begin{figure*}[t]
\centering  
{\includegraphics[width=0.95\linewidth]{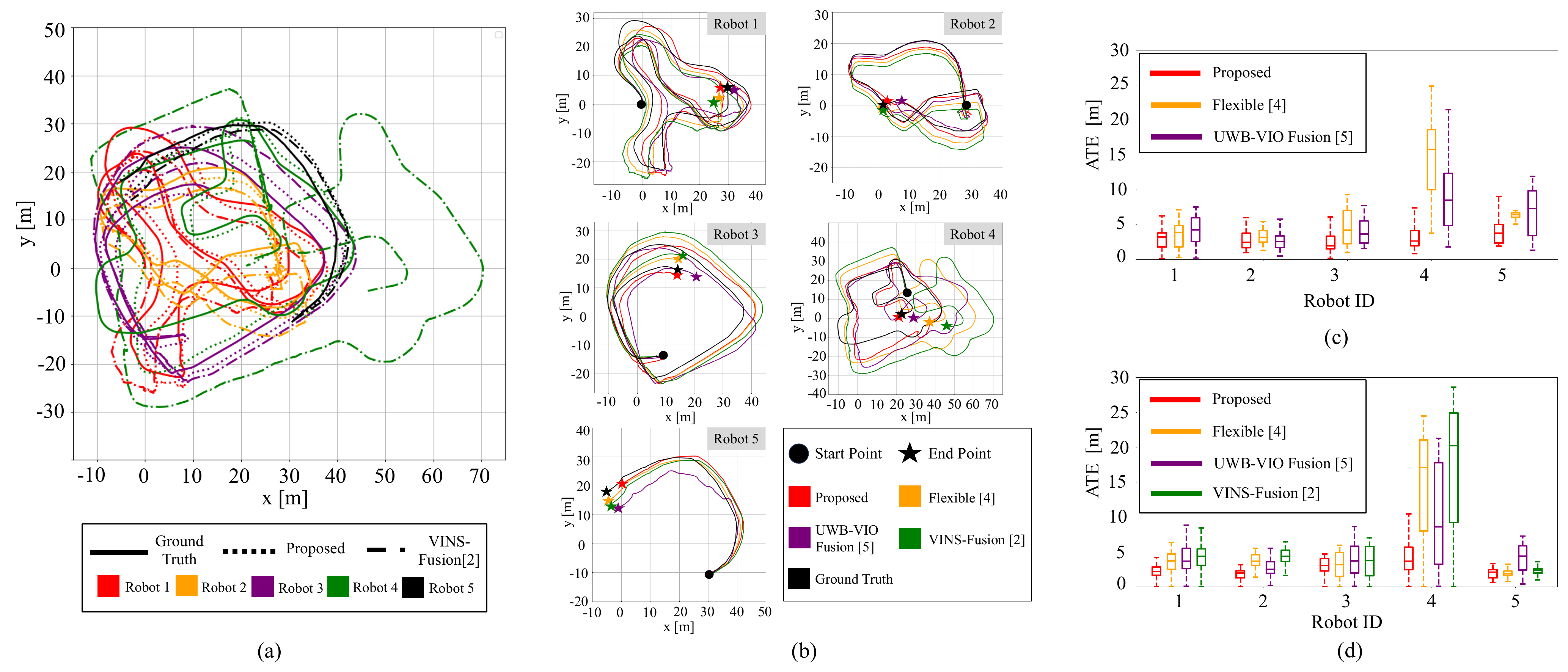}}
\captionsetup{font=footnotesize}
\caption{The overall results of Sequence~1. (a) The top view of the multi-robot trajectories of the proposed, VINS-Fusion\cite{single_UAV_odom1} with known initial pose, and ground truth. The initial positions of the robots are marked with colored circles. (b) The top view of the trajectories of each robot. (c) ATE comparison of the proposed, Flexible\cite{multi-UAV_local_3_nguyen2021flexible}, and UWB-VIO Fusion\cite{multi-UAV_local_15_9896952} without known initial pose. (d) ATE comparison of the proposed, Flexible\cite{multi-UAV_local_3_nguyen2021flexible}, UWB-VIO Fusion\cite{multi-UAV_local_15_9896952}, and VINS-Fusion\cite{single_UAV_odom1} with known initial pose.}
\label{result exp1}
\end{figure*}

\begin{figure*}[t]
\centering  
{\includegraphics[width=0.95\linewidth]{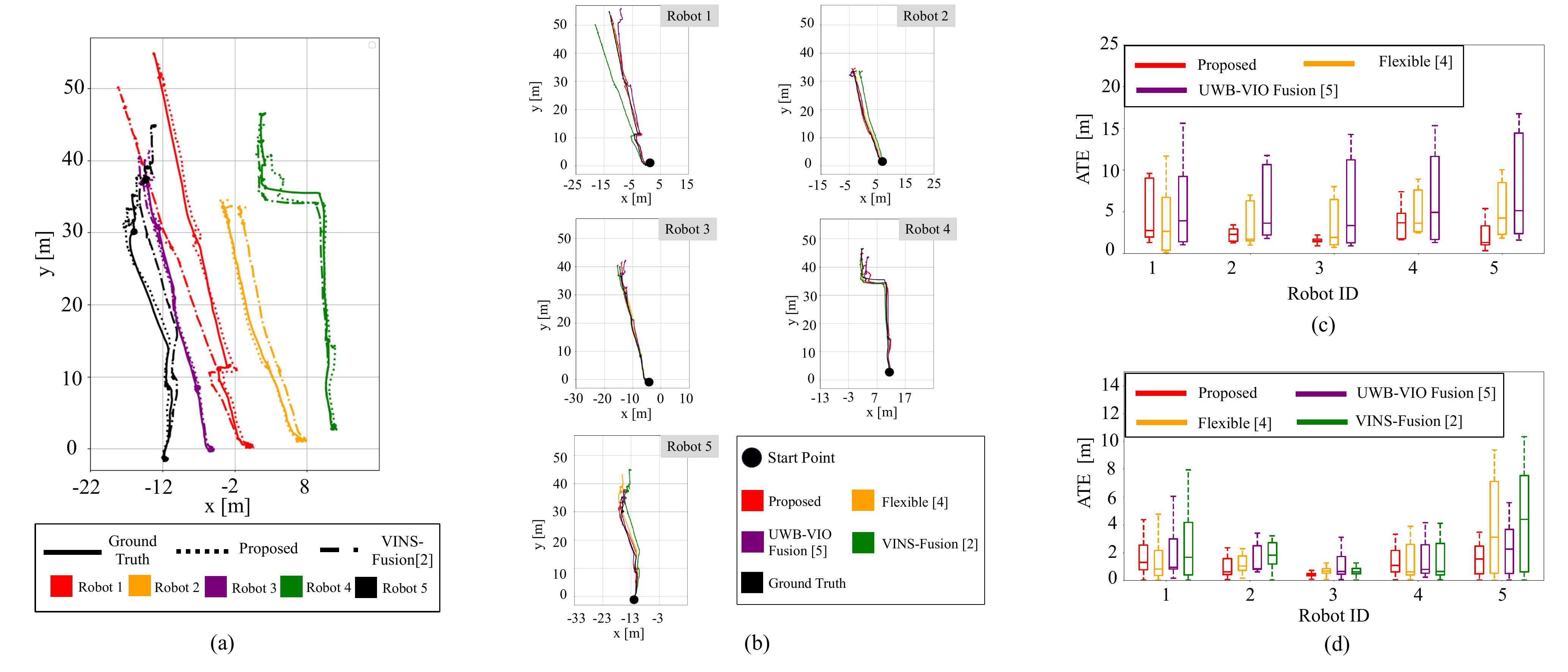}}
\captionsetup{font=footnotesize}
\caption{The overall results of Sequence~2. (a) The top view of the multi-robot trajectories of the proposed, VINS-Fusion\cite{single_UAV_odom1} with known initial pose, and ground truth. The initial positions of the robots are marked with colored circles. (b) The top view of the trajectories of each robot. (c) ATE comparison of the proposed, Flexible\cite{multi-UAV_local_3_nguyen2021flexible}, and UWB-VIO Fusion\cite{multi-UAV_local_15_9896952} without known initial pose. (d) ATE comparison of the proposed, Flexible\cite{multi-UAV_local_3_nguyen2021flexible}, UWB-VIO Fusion\cite{multi-UAV_local_15_9896952}, and VINS-Fusion\cite{single_UAV_odom1} with known initial pose.}
\label{result exp2}
\end{figure*}

The overall results of Sequence~1 are shown in \figref{result exp1} and Sequence~2 in \figref{result exp2}. Unlike the previous related works, all the robots moved a relatively long distance, and the average lengths of the trajectories were 242.04\,m and 75.89\,m for Sequences~1 and 2, respectively. The length of each robot trajectory and the RMSE of absolute trajectory error are shown in Table~\Romannumeral1. The average RMSE values of the absolute trajectory error (ATE) were decreased in all trajectories by our proposed method. In particular, even when the local VIO included a large scale error, such as Robot 4 in Sequence~1, the proposed method estimated the position well compared with other state-of-the-art methods.

\begin{table*}[h]
\captionsetup{font=footnotesize}
\centering
\caption{Comparison on the Sequence~1 and Sequence~2. All errors are represented in the form of RMSE of absolute trajectory error [m].}
\label{result table 1}
{\scriptsize
\begin{tabular}{c|c|c|ccc|cccc}
\hline
\multicolumn{3}{c|}{} & \multicolumn{3}{c|}{RMSE (initial pose unknown)} & \multicolumn{4}{c}{RMSE (initial pose known)} \\ \hline\hline
& Robot & Length of & \multirow{2}{*}{Ours} & \multirow{2}{*}{Flexible\cite{multi-UAV_local_3_nguyen2021flexible}} & UWB-VIO & \multirow{2}{*}{Ours} & \multirow{2}{*}{Flexible\cite{multi-UAV_local_3_nguyen2021flexible}} & UWB-VIO & \multirow{2}{*}{VINS-Fusion\cite{single_UAV_odom1}}\\
& ID & trajectory (m) & & & Fusion\cite{multi-UAV_local_15_9896952} & & & Fusion\cite{multi-UAV_local_15_9896952} & \\ \hline
\multirow{6}{*}{\rotatebox{90}{Sequence 1}}& 1 & 265.41 & \textbf{3.40} & 3.86 & 4.63 & \textbf{2.77} & 3.73 & 4.61 & 4.61 \\
& 2 & 165.50 & \textbf{3.19} & 3.44 & 3.23 & \textbf{1.95}& 3.84 & 3.05 & 4.47 \\
& 3 & 449.16 & \textbf{4.16} & 5.45 & 4.34 & \textbf{3.19} & 3.58 & 4.52 & 4.22 \\
& 4 & 246.10 & \textbf{3.51} & 15.78 & 10.83 & \textbf{6.28} & 16.46 & 11.93 & 19.50 \\
& 5 & 84.05 & \textbf{4.87} & 6.30  &  7.65& \textbf{2.00} & 2.13 & 4.59 & 2.54 \\ \cline{2-10}
& Avg. & 242.04 & \textbf{3.83} & 6.97 & 6.14 & \textbf{3.24} &5.94 & 5.74 & 7.07 \\ \hline
\hline
\multirow{6}{*}{\rotatebox{90}{Sequence 2}}& 1 & 81.72 & \textbf{4.18} & 5.77 & 7.20 & 2.21 & \textbf{2.10} & 2.52 & 3.73 \\
& 2 & 61.89 & \textbf{2.18} & 4.28 & 6.97& \textbf{1.12} & 1.27 & 1.75 & 2.07 \\
& 3 & 70.93 & \textbf{1.64} & 4.39 & 7.30 & \textbf{0.53} & 0.68 & 1.29 & 0.68 \\
& 4 & 88.64 & \textbf{3.03} & 5.50 & 7.92 & \textbf{1.69} & 1.83 & 1.96 & 1.95 \\
& 5 & 76.28 & \textbf{2.43} & 6.21 & 10.15 & \textbf{1.79} & 5.10 & 2.75 & 5.61 \\ \cline{2-10}
& Avg. & 75.89 & \textbf{2.69} & 5.23  & 7.91 & \textbf{1.47} & 2.20 & 2.05  & 2.81  \\ \hline
\end{tabular} 
}
\end{table*}

\section{CONCLUSIONS}
In summary, we proposed a novel multi-robot localization method that is robust to the long-term drift error by using VIO and UWB range data together. Moreover, the proposed method showed notable improvements in the accuracy of the multi-robot pose estimation over existing approaches. As a note, the cumulative error and scale error of conventional VIO were reduced by integrating UWB range data and local VIO in a complementary fashion through a hierarchical and adaptive optimization framework. The robustness and effectiveness of the proposed method were verified using real-world experiments with the multi-robot systems consisting of UAVs and UGVs on large outdoor environment. The experimental results showed that the proposed method reduced both long-term drift that was caused by long trajectory, and scale error that occurred in one of the robots in the group. The comparison methods also filter the measurements and consider global consistency and VIO characteristics. However, since the entire measurements are ultimately used in the optimization phase, scale error or drift affects the overall multi-robot localization. The proposed method leverages the advantages of the UWB sensor and VIO to complement these problems. The relative position structure is fixed using only range measurement, then the drift and scale error of each VIO is calculated and the poses are corrected. As future works, we will improve the proposed method to operate robustly in complex environments where dynamic obstacles exist, and expand it to a collaborative mapping scenario.

\bibliographystyle{IEEEtran}
\bibliography{references}

\end{document}